\documentclass{article}

\usepackage{xspace}
\usepackage{wrapfig}
\usepackage{lipsum}
\usepackage{subcaption}
\usepackage{caption}
\usepackage{multirow}
\usepackage{diagbox}
\usepackage{array}
\usepackage{supertabular}
\usepackage{multicol}
\usepackage{amsmath}
\DeclareMathOperator*{\argmax}{argmax}
\usepackage{bm}
\usepackage{dsfont}
\usepackage{enumitem}

\usepackage{graphicx}

\newcommand{\fasterrcnn}{\textit{Faster R-CNN}\xspace}
\newcommand{\fasterrcnnv}{\textit{Faster R-CNN\_v2}\xspace}
\newcommand{\retinanet}{\textit{RetinaNet}\xspace}
\newcommand{\retinanetv}{\textit{RetinaNet\_v2}\xspace}
\newcommand{\ssd}{\textit{SSD300}\xspace}
\newcommand{\ssdlite}{\textit{SSDLite}\xspace}

\newcommand{\origimg}{\textit{origImg}\xspace}
\newcommand{\dalleimg}{\textit{dalleImg}\xspace}
\newcommand{\stableimg}{\textit{stableImg}\xspace}
\newcommand{\advorigimg}{\textit{advOrigImg}\xspace}
\newcommand{\advdalleimg}{\textit{advDalleImg}\xspace}
\newcommand{\advstableimg}{\textit{advStableImg}\xspace}

\newcommand{\uqms}{\textit{UQM}\xspace}

\newcommand{\map}{\textit{mAP}\xspace}

\newcommand{\vr}{\textit{VR}\xspace}
\newcommand{\se}{\textit{SE}\xspace}
\newcommand{\mi}{\textit{MI}\xspace}
\newcommand{\tv}{\textit{TV}\xspace}
\newcommand{\ps}{\textit{PS}\xspace}

\newcommand{\rs}{\textit{RS}\xspace}

\newcommand{\sdm}{\textit{SDM}\xspace}

\newcommand{\rsmap}{$RS_{mAP}$\xspace}
\newcommand{\rsuqm}{$RS_{uq}$\xspace}

\newcommand{\dalle}{\textit{DALL-E-3}\xspace}
\newcommand{\sd}{\textit{Stable Diffusion-3}\xspace}

\usepackage{xcolor}  
\newcommand{\conclusion}[1]{
\begin{center}
    \fcolorbox{black}{gray!10}{\parbox{.97\columnwidth}{#1}}
\end{center}
}

\usepackage{PRIMEarxiv}

\usepackage[utf8]{inputenc} 
\usepackage[T1]{fontenc}    
\usepackage{hyperref}       
\usepackage{url}            
\usepackage{booktabs}       
\usepackage{amsfonts}       
\usepackage{nicefrac}       
\usepackage{microtype}      
\usepackage{lipsum}
\usepackage{fancyhdr}       
\usepackage{graphicx}       
\graphicspath{{media/}}     

\title{Assessing the Uncertainty and Robustness of the Laptop Refurbishing Software
}

\author{
Chengjie~Lu
\\
Simula Research Laboratory and
University of Oslo\\
Oslo, Norway \\
\texttt{chengjielu@simula.no} \\
\And
 Jiahui~Wu
\\
Simula Research Laboratory and
University of Oslo\\
Oslo, Norway \\
\texttt{jiahui@simula.no} \\
\And
 Shaukat~Ali
\\
Simula Research Laboratory\\
Oslo, Norway \\
\texttt{shaukat@simula.no} \\
\And
Mikkel~Labori~Olsen
\\
Danish Technological Institute\\
Odense, Denmark \\
\texttt{miol@teknologisk.dk} \\
}

\begin{document}
\maketitle

\begin{abstract}
Refurbishing laptops extends their lives while contributing to reducing electronic waste, which promotes building a sustainable future. To this end, the Danish Technological Institute (DTI) focuses on the research and development of several robotic applications empowered with software, including laptop refurbishing. 
Cleaning represents a major step in refurbishing and involves identifying and removing stickers from laptop surfaces. 
Software plays a crucial role in the cleaning process. For instance, the software integrates various object detection models to identify and remove stickers from laptops automatically. However, given the diversity in types of stickers (e.g., shapes, colors, locations), identification of the stickers is highly uncertain, thereby requiring explicit quantification of uncertainty associated with the identified stickers. Such uncertainty quantification can help reduce risks in removing stickers, which, for example, could otherwise result in software faults damaging laptop surfaces. For uncertainty quantification, we adopted the Monte Carlo Dropout method to evaluate six sticker detection models (\sdm{s}) from DTI using three datasets: the original image dataset from DTI and two datasets generated with vision language models, i.e., \dalle and \sd. In addition, we presented novel robustness metrics concerning detection accuracy and uncertainty to assess the robustness of the \sdm{s} based on adversarial datasets generated from the three datasets using a dense adversary method.

Our evaluation results show that different \sdm{s} perform differently regarding different metrics. Based on the results, we provide \sdm selection guidelines and lessons learned from various perspectives.
\end{abstract}


\section{Introduction}
The European Union's Circular Economy Action Plan (CEAP) highlights the need for sustainable operations to promote circular economy processes and encourage sustainable consumption~\cite{doi/10.2779/05068}.
One essential activity is refurbishing electronic devices, e.g., laptops, to extend their lives, reduce electronic waste, and provide affordable options for consumers. A critical and time-consuming step in refurbishment is removing stickers from the laptop by first identifying stickers and their locations. 
Manual cleaning is time-consuming and faces challenges in finding enough skilled workers, and current automation solutions are not built for this amount of variation, limiting sustainability and scalability.
Thus, novel solutions to automate the processes are needed. Robotics offers a promising solution to simplify and scale up this process, increasing efficiency and reducing labor costs.

The Danish Technological Institute (DTI) develops, applies, and transfers technology to industry and society. One leading area that DTI focuses on is laptop refurbishing automatically with robots, where software is a crucial part of all laptop refurbishing steps. The software responsible for the cleaning process in laptop refurbishment integrates deep neural networks (DNN)--based sticker detection models (\sdm{s}) built by DTI for automatic sticker detection, which is the basis for successful automatic sticker removal. The \sdm{s} are built on open-source object detection DNNs and trained using a sticker detection dataset specially designed by DTI.
Due to inappropriate model architecture or insufficient training data, the design and training process may introduce uncertainties into the \sdm{s}, making them vulnerable under certain conditions, such as adversarial attacks, noisy data, or unforeseen input distributions~\cite{carlini2017towards}. This vulnerability highlights the need to quantify uncertainty and evaluate the robustness of the \sdm{s} and software they are integrated in, which is crucial for trustworthy sticker removal, as incorrect detection may damage the laptop surface or incomplete sticker removal.

As a first step towards holistic uncertainty quantification (UQ) and handling in laptop refurbishing robotic software, we conduct a comprehensive empirical evaluation to assess the \sdm{s} DTI uses regarding detection accuracy, prediction uncertainty, and adversarial robustness.
Specifically, we adopt Monte Carlo Dropout (MC-Dropout)~\cite{gal2016dropout} as the UQ method to capture the uncertainty in model predictions. 
We run the model to perform multiple predictions, and based on these predictions, we calculate two types of UQ metrics: label classification UQ metrics and bounding box regression UQ metrics. 
We employ Dense Adversary Generation (DAG)~\cite{xie2017adversarial} as the adversarial attack and define \textit{robustness score (\rs)} to measure the robustness of the \sdm{s}. \rs measures robustness from two perspectives: robustness concerning predictive precision and robustness concerning prediction uncertainty. Regarding benchmark datasets, we construct benchmark datasets from three data sources: datasets provided by our partner DTI, datasets synthesized by prompting two vision language models (VLMs), i.e., \dalle~\cite{betker2023improving} and \sd~\cite{rombach2021highresolution}, and datasets created using the adversarial attack, i.e., DAG. Our evaluation results show that different \sdm{s} achieve different performance regarding different evaluation metrics. Specifically, regarding sticker detection accuracy, \fasterrcnnv is recommended as the best \sdm, while \retinanetv achieved the overall best performance regarding prediction uncertainty. Regarding adversarial robustness, \fasterrcnnv and \retinanetv are recommended as the best \sdm{s}. We also provide guidelines for \sdm selection and lessons learned.

In summary, our contributions are: 1) we conduct an empirical study to evaluate the detection accuracy, uncertainty, and robustness of six \sdm{s} implemented in robotic software to detect stickers on the laptop; 2) we employ MC-Dropout as the UQ method and calculate UQ metrics to measure uncertainty in classification and regression; 3) we present a novel robustness metric, i.e., \rs, to measure the robustness by considering model uncertainty; 4) we provide guidelines for \sdm selection from different perspectives.

\section{Industrial Context \& Background} \label{sec:industrialcontext}

\subsection{Industrial Context} \label{subsec:industrial}
DTI is a leading Danish institute focused on developing advanced technology and transferring the technology to industry. A key area that DTI focuses on is robotics. Within robotics, DTI is leading an initiative for robotic applications for sustainable operations deemed critical in CEAP.
Adopting circular economy principles from the CEAP could substantially boost the EU's GDP by 0.5\% by 2030, creating approximately 700,000 new jobs~\cite{doi/10.2779/05068}. Electrical and electronic equipment, one of the fastest-growing waste streams in the EU, currently sees less than 40\% recycled, resulting in significant value loss~\cite{doi/10.2779/05068}.
This scenario underscores the need to refurbish electronics such as laptops, which is the context of this paper. A typical refurbishment process consists of disassembly, cleaning, inspection, restoration, reassembly, and testing. Collaborative robots performing these steps often require human involvement for challenging tasks, e.g., non-trivial disassembly tasks.
In particular, an essential and time-consuming step in the refurbishment process is cleaning the stickers on the laptop, which can be further divided into two tasks: sticker detection and removal. 
Accurate detection of stickers with high confidence is crucial, as incorrect detection may result in damage to the laptop surface or incomplete sticker removal. 
Although the sticker detection task is straightforward for humans, the unlimited number of stickers and ever-changing types of laptops pose a significant challenge for automating and scaling this process. This issue of handling extreme variability is common across many loops of the circular economy, including recycling, refurbishing, repairing, and remanufacturing. Thus, having a method to manage this variability in a scalable and automated way is crucial for becoming more circular.

Robotics and AI systems can automate the sticker detection process. To this end, DTI integrates DNN-based \sdm{s} in their robotic software to automatically identify laptop stickers. During the DNN designing and training process, uncertainty might arise due to inappropriately designed model architectures or insufficient training data. For example, some laptop brands may not be included in the training data, so the \sdm{s} may not detect stickers correctly or behave highly uncertainly (e.g., detecting the same object multiple times but giving different detection results). Therefore, we must holistically quantify and manage the uncertainty in the entire refurbishment process implemented as software inside robots from different perspectives, such as environmental uncertainties, machine learning models employed, and humans interacting with robots. The context of this work is the software responsible for laptop sticker detection that integrates \sdm{s}. Quantifying uncertainty in \sdm{s} is crucial from several perspectives. First, it ensures safe and reliable decisions of robots, especially when robots work alongside humans. Besides, managing detection uncertainty can significantly improve quality control, reduce errors, and increase accuracy. In addition, adaptability can be enhanced by helping the model adapt to different sticker types and conditions, effectively guiding human intervention when needed. Finally, workflow efficiency can be optimized by prioritizing cases that require human attention, thereby reducing downtime and streamlining workflows.

\subsection{Bayesian Uncertainty Quantification in Deep Learning}\label{subsec:uq_dl}
Bayesian Neural Networks (BNNs) provide a probabilistic approach to neural network modeling that allows UQ through Bayesian inference~\cite{tran2019bayesian}. In BNNs, model weights are treated as random variables with a prior distribution reflecting the initial belief about the weights before observing any data. A common choice is a Gaussian distribution: $p(\mathbf{w}) = \mathcal{N}(\mathbf{w} \mid \mathbf{0}, \mathbf{I})$. Given a dataset $\mathbf{X}=\{x_1, x_2, ..., x_n\}$, $\mathbf{Y}=\{y_1, y_2, ..., y_n\}$, Bayesian inference computes the posterior over the weights as: $p(\mathbf{w}\mid\mathbf{X}, \mathbf{Y})$. Instead of providing a single output prediction, BNNs produce a distribution of possible outputs that captures the uncertainty about the prediction. The posterior combines the prior and the likelihood of the observed data, which is often difficult to compute exactly because the marginal probability $p(\mathbf{Y}\mid\mathbf{X})$ required to evaluate the posterior cannot be evaluated analytically~\cite{mackay1992practical}. Hence, various Bayesian approximation methods have been proposed to perform Bayesian inference.

Gal and Ghahramani~\cite{gal2016dropout} proposed MC-Dropout as a Bayesian approximation method, which shows that introducing dropout in DNN can be interpreted as the approximation of a probabilistic Bayesian model in deep Gaussian processes. Dropout is commonly applied in deep learning models to avoid over-fitting by randomly dropping units along with their connections from the neural network~\cite{srivastava2014dropout}. Thus, the predictive distribution is approximated by sampling multiple predictions:
\vspace{-8pt}
\begin{equation}
\vspace{-6pt}
    p(y^{*} \mid x^{*}, \mathbf{X}, \mathbf{Y}) \approx \frac{1}{T}p(y^{*} \mid x^{*}, \mathbf{w}^{'}),
\vspace{3pt}
\end{equation}
where $x^*$ is the input sample, $y^*$ is the predicted output, $T$ is the number of predictions, and $\mathbf{w}^{'}$ is the model weights after dropout. The $T$ DNN predictions are different because neurons are randomly dropped in each prediction, based on which we quantify the uncertainty of the overall predictions.

\section{Methodology}

\begin{figure}[!htbp]
    \centering
    \resizebox{0.8\columnwidth}{!}{\includegraphics{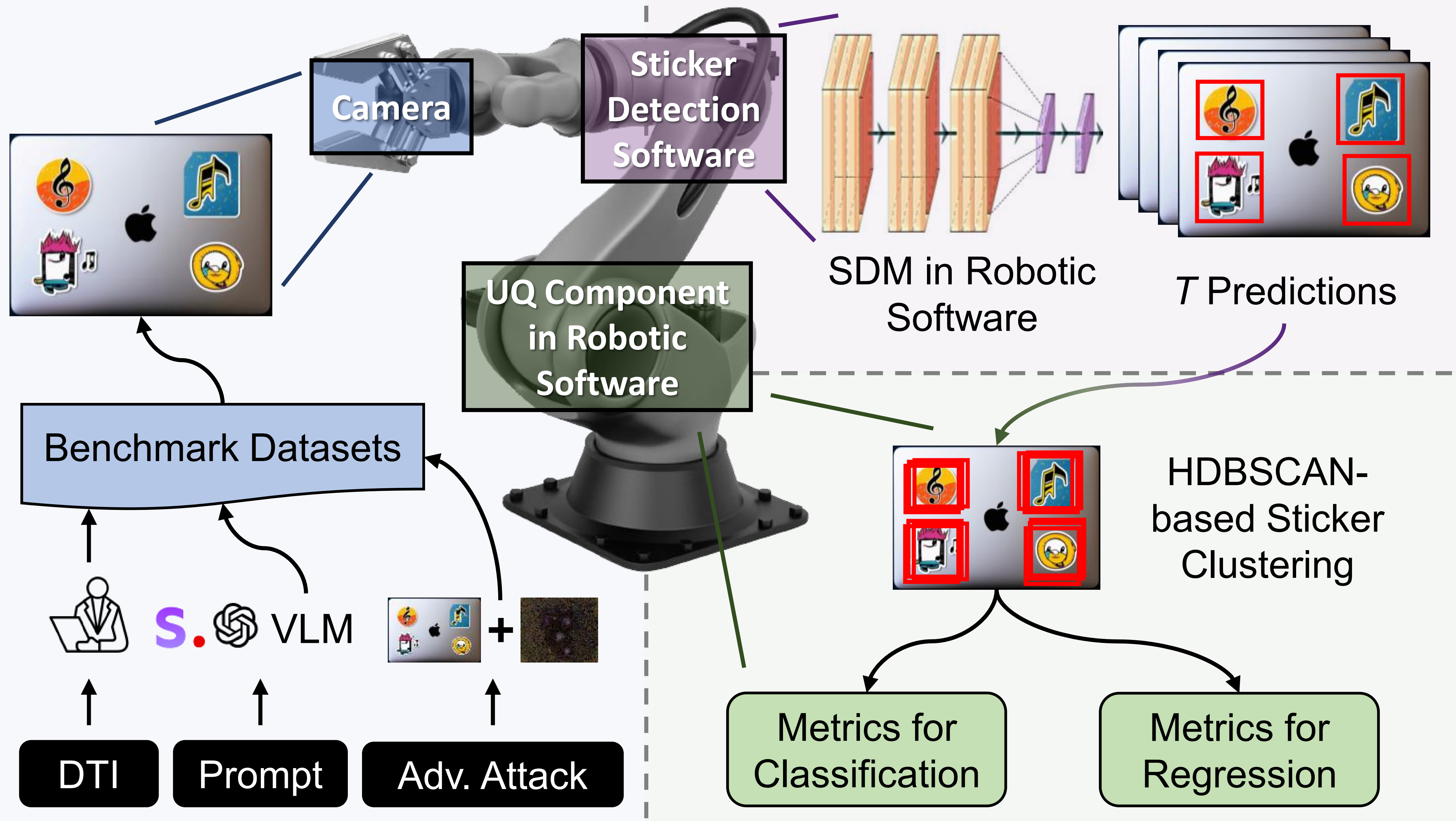}}
    \caption{Overview of the Uncertainty Quantification Process for Sticker Detection Models in the Robotic Software.}
    \label{fig:overview}
    \vspace{-10pt}
\end{figure}

Figure~\ref{fig:overview} presents an overview of the UQ process for the \sdm{s} employed in the robotic software for laptop refurbishing. 
As the figure shows, the UQ process starts with creating benchmark datasets from three data sources: datasets from DTI, datasets synthesized by prompting two VLMs, and datasets created using an adversarial attack technique. For each image in the benchmark datasets, the robot's camera captures the image, which is then processed by specialized robotic software for sticker detection. Specifically, the \sdm in the robotic software makes \textit{T} predictions based on the MC-Dropout method to capture the prediction uncertainty. The \textit{T} predictions are then fed to the UQ component in the robotic software to calculate UQ metrics. The UQ component first employs the density-based algorithm, HDBSCAN, to cluster the detected stickers in \textit{T} predictions and then calculates the UQ metrics based on the detected stickers.
We consider two types of UQ metrics: label classification UQ metrics and bounding box regression UQ metrics. We present the MC-Dropout method in Section~\ref{subsec:dropout} and the UQ metrics in Section~\ref{subsec:uqmetrics}. Sections~\ref{subsec:llms} and~\ref{subsec:adv} introduce VLM-based and adversarial attack-based dataset creation.

\subsection{Monte Carlo Dropout-Based Uncertainty Quantification}\label{subsec:dropout}
Bayesian probabilistic theory~\cite{bernardo2009bayesian} is a universal theoretical framework for uncertainty reasoning, and as discussed in Section~\ref{subsec:uq_dl}, MC-dropout is a Bayesian approximation for quantifying uncertainty. Hence, we adopt MC-Dropout as the UQ method and inject inference-time activated dropout layers into the pre-trained \sdm{s}. By doing so, we build uncertainty-aware variants of the original models. Specifically, for a \sdm $f$, we first build its variant as $f^p$ by applying dropout layers with dropout rate $p$. Next, we let $f^p$ make $T$ predictions for a given image input $X$ to obtain a sample of outputs $O=\{f^p_t\}_{t=1}^T$.
The outputs $O$ represent a sample from the model's predictive distribution, and MC-Dropout quantifies the uncertainty of $f^p$'s outputs by extracting information regarding the variability of the $T$ predictions. We set $T$ to 20 according to existing guidelines~\cite{gal2016dropout}. \looseness=-1

\subsection{Uncertainty Quantification Metrics}\label{subsec:uqmetrics}

\subsubsection{Basic Concepts}
As discussed in Section~\ref{subsec:dropout}, we utilize MC-Dropout to quantify the uncertainty over $T$ prediction outputs given an input image $X$: $O=\{f^p_t\}_{t=1}^T$. A prediction $f^p_t$ is a vector containing $N$ different detected objects, and a detected object $f^p_t(i), (1 \leq i \leq N)$ is characterized by its label classification and bounding box regression.
The label classification assigns the softmax score $sm_{f^p_t(i)}$ for the candidate objects of three classes: stickers, logos, and backgrounds. The softmax score is a vector of probabilities, and we take the position with the maximum probability value as the predicted label, i.e., $argmax()$. The bounding box regression returns four different real values $box_{f^p_t(i)}=\{x_1, x_2, y_1, y_2\}$ to locate the detected object. Specifically, a bounding box is a rectangle whose start and end points are denoted as $(x_1, y_1)$ and $(x_2, y_2)$. In a deterministic model without uncertainty, all predictions should be the same, for example, a detected object $f^p_t(i)$ in prediction $f^p_t$ should also be detected in other predictions and have the same softmax score and bounding box regression values as those in other predictions. However, recall that MC-Dropout quantifies the uncertainty by utilizing $T$ model predictions with randomly dropped units. Therefore, each prediction is not necessarily the same. The model based on MC-Dropout can produce predictions with different locations for each object in a single image. To this end, a clustering method is needed to cluster objects in the $T$ predictions. Details will be introduced in the following section.

\subsubsection{HDBSCAN-based Object Clustering} We adopt the hierarchical density-based clustering algorithm, HDBSCAN~\cite{campello2013density}, to cluster objects based on their predicted bounding boxes. 
The algorithm identifies clusters by calculating core distances for each point and adjusting these distances based on local density. It then builds a minimum spanning tree to form a hierarchy of clusters, which is condensed to highlight meaningful clusters. This hierarchical structure enables the recognition of clusters of various shapes and sizes. 
Two key parameters need to be adjusted: \textit{minimum samples ($minSamples$)}, which sets the minimum number of samples in the neighborhood for a point to be considered a core point, and \textit{minimum cluster size ($minPts$)}, which determines the minimum number of points to be considered as a cluster. Based on our preliminary study, we set both $minSamples$ and $minPts$ to 3.
As a result, each cluster represents a detected object in the $T$ predictions whose label classification softmax score and the bounding box regression are denoted as $\{sm_k\}_{k=1}^{W}$ and $\{box_k\}_{k=1}^{W}$. Note that $W$ is the number of times the object is detected, and it is in the range $[1, T]$, as some objects are not always detected in all $T$ predictions. We then calculate UQ metrics for each detected object in the $T$ predictions.

\subsubsection{Uncertainty Metrics for Label Classification} We employ three common metrics to measure uncertainty in classification tasks~\cite{gal2016uncertainty}, including Variation Ratio (\vr)~\cite{freeman1965elementary}, Shannon Entropy (\se)~\cite{shannon1948mathematical}, and Mutual Information (\mi)~\cite{shannon1948mathematical}. \vr is a measure of dispersion. It is calculated by determining the proportion of cases that are not in the mode (the most frequent category) relative to the total number of cases. For the detected object in each cluster, we first extract a set of labels as the classes with the highest softmax scores:
\vspace{-3pt}
\begin{equation}
\vspace{-1pt}
    \mathbf{y}=\{\argmax(sm_k)\}_{k=1}^W,
\vspace{-2pt}
\end{equation}
we then find the mode of the distribution as:
\vspace{-1pt}
\begin{equation}
\vspace{-1pt}
    c^* = \argmax_{c=1,...,C}\sum_k\mathds{1}[\mathbf{y}^k=c],
\vspace{-1pt}
\end{equation}
where $\mathds{1}$ is the indicator function. The number of times $c^*$ was sampled is $\sum_k\mathds{1}[\mathbf{y}^k=c^*]$. Finally, \vr is calculated as:
\vspace{-1pt}
\begin{equation}
\vspace{-1pt}
    \vr=1-\frac{\sum_k\mathds{1}[\mathbf{y}^k=c^*]}{W}.
\end{equation}
\vr ranges from 0 to $2/3$. It attains the maximum value of $2/3$ when all three classes are sampled equally and the minimum value of 0 when only a single class is sampled.

\se captures the average amount of information contained in the predictive distribution. For a detected object, we calculate \se by considering softmax scores over $T$ predictions, i.e., $\{sm_k\}_{k=1}^W$ as:
\vspace{-1pt}
\begin{equation}
\vspace{-1pt}
    SE=-\sum_{c=1}^{nc}(\frac{1}{W}\sum_{k=1}^{W}sm_k(c)) \times \log (\frac{1}{W}\sum_{k=1}^{W}sm_k(c)),
\end{equation}
where $nc$ denotes the number of classes, i.e., 3. \se values range from 0 to $\log(3)$, reaching a maximum of $\log(3)$ when all classes are predicted to have the same softmax score, i.e., $1/3$, indicating the most uncertain case. Its minimum value is 0 when the probability of one class is 1 and all other probabilities are 0, indicating no uncertainty in the classification.

\mi quantifies the information difference between the predicted and posterior of the model parameters, providing a different uncertainty measure for classification tasks. For the detected object in each cluster, \mi can be computed as~\cite{gal2016uncertainty}:
\vspace{-1pt}
\begin{equation}
\vspace{-1pt}
    \begin{split}
        \mi=-\sum_{c=1}^{nc}(\frac{1}{W}\sum_{k=1}^{W}sm_k(c)) \times \log (\frac{1}{W}\sum_{k=1}^{W}sm_k(c)) \\
        + \frac{1}{W}\sum_{k=1}^{W}\sum_{c=1}^{nc}sm_k(c) \times \log sm_k(c),
    \end{split}
\end{equation}
where $nc$ denotes the number of classes, i.e., 3. \mi measures the model's confidence in its outputs. It ranges from 0 to 1, and the larger the \mi value, the higher the uncertainty.

\subsubsection{Uncertainty Metrics for Bounding Box Regression} For each detected object, we estimate the uncertainty for bounding box regression using Total Variance (\tv)~\cite{feng2018towards} and Predictive Surface (\ps)~\cite{catak2021prediction}. \tv captures the uncertainty in the bounding box regression by calculating the trace of the covariance matrix of $\{box_k\}_{k=1}^W$, which sums the variances of each variable in the bounding box. Specifically, a bounding box has four variables used to locate an object: $box=\{x_1, x_2, y_1, y_2\}$, for variable $v\in box$, we calculate its variance as:
\vspace{-1pt}
\begin{equation}
\vspace{-1pt}
    \delta_v^2=\frac{1}{W-1}\sum_{k=1}^{W}(v_k-\mu_v)^2,
\end{equation}
where $\delta^2$ is the variance of $v$, $\mu$ is the mean value of $v$, and $W$ is the number of times the object is detected. We then calculate \tv by summing the variances for all variables:
\vspace{-1pt}
\begin{equation}
\vspace{-1pt}
    \tv=\sum_{v\in box}\delta_v^2.
\end{equation}
\tv ranges in $[0, +\infty)$, and larger \tv means higher uncertainty.

\ps was originally proposed to quantify prediction uncertainty in object detection models for autonomous driving~\cite{catak2021prediction}. It measures uncertainty by considering the convex hull of each corner point of the predicted bounding box. A convex hull is the smallest convex shape enclosing all points and \ps estimates uncertainty by calculating the area covered by the convex hull. To be concrete, four corner points define the bounding box to locate an object: $(x_1, y_1)$, $(x_2, y_2)$, $(x_2, y_1)$, and $(x_1, y_2)$. For the object in each cluster where the bounding box is detected $W$ times, i.e., $\{box_k\}_{k=1}^W$, we first obtain the cluster of each corner point as:
\vspace{-1pt}
\begin{equation}
\vspace{-1pt}
    \begin{split}
        CP=\{(x_1, y_1)_k, (x_2, y_2)_k, (x_2, y_1)_k, (x_1, y_2)_k\}_{k=1}^W,
    \end{split}
\end{equation}
we then identify the convex hulls for its four corner points, respectively, as:
\vspace{-1pt}
\begin{equation}
\vspace{-1pt}
    convexh=ConvexHull(cp), cp \in CP.
\end{equation}
Finally, we calculate \ps by averaging the area covered by the convex hulls of the four corner points to approximate prediction uncertainty:
\vspace{-1pt}
\begin{equation}
\vspace{-1pt}
    \ps=\frac{1}{|CP|}\sum_{cp\in CP}Area(convexh_{cp}).
\end{equation}
\ps ranges in $[0, +\infty)$, and a higher \ps value indicates higher variability of the predicted corner points, meaning higher uncertainty in the bounding box prediction.

\begin{figure}[!t]
    \centering
    \resizebox{0.8\columnwidth}{!}{\includegraphics[width=\linewidth]{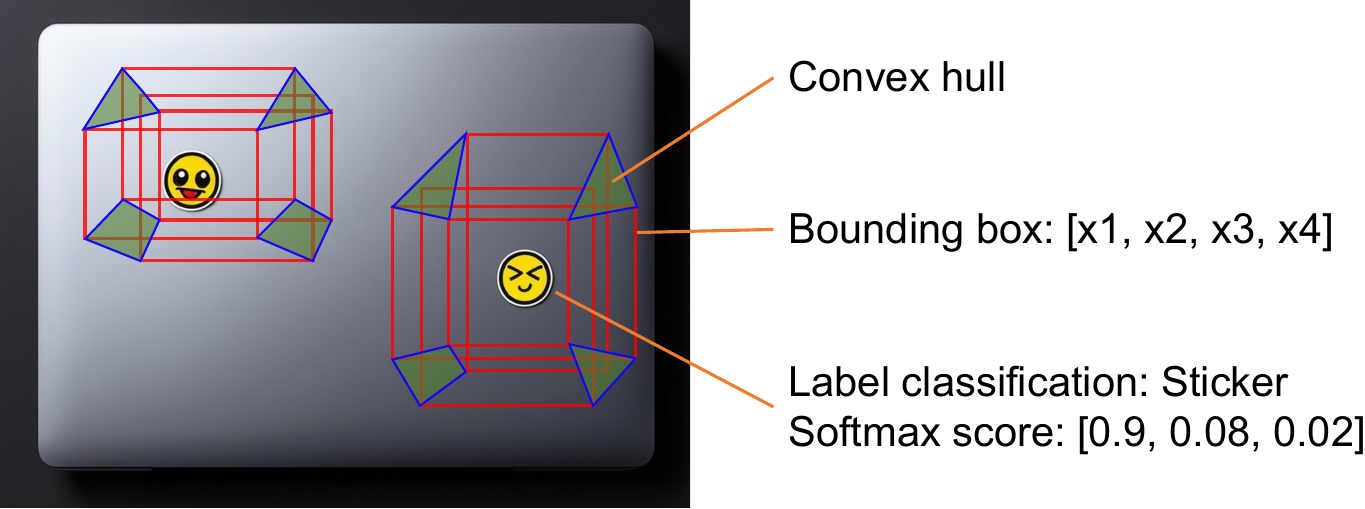}}
    \caption{An example for sticker detection. 
    The left column shows the laptop image with two stickers on it. The right column shows the sticker detection outputs and what the convex hull looks like in the left image. 
    }
    \label{fig:adv}
    \vspace{-6pt}
\end{figure}

Recall that for an input image $X$, multiple objects will be detected, so after performing the HDBSCAN clustering algorithm, multiple clusters will be obtained, each cluster representing a detected object. To get the uncertainty of the input image $X$, for each UQ metric, we take the average of all metric values of all detected objects in $X$. Figure~\ref{fig:adv} shows an example of using the \sdm to detect stickers on a laptop and how to calculate UQ metrics based on the model outputs. The \sdm makes \textit{T} predictions, and each prediction returns a label based on the softmax score and a bounding box containing four corner points. The convex hull of each corner point is then calculated. The UQ metrics for label classification are calculated based on the softmax scores in \textit{T} predictions. Regarding UQ metrics for bounding box regression, \tv is calculated based on the variance of each corner point, and \ps is calculated based on the area covered by convex hulls.

\subsection{VLM-based Image Generation}\label{subsec:llms}
VLMs are generic pre-trained multimodal models that learn from large-scale image-text pairs sourced from the Internet, enabling them to directly address downstream visual tasks without task-specific fine-tuning~\cite{zhang2024vision}. Therefore, we apply the zero-shot generalization ability of VLMs to generate a synthetic image dataset to study whether \sdm{s} can identify stickers on laptops that they have not encountered before. Following the representative categories of VLMs, i.e., autoregressive and diffusion models~\cite{li2023gligen}, we select two state-of-the-art VLMs: \dalle~\cite{betker2023improving} and \sd~\cite{rombach2021highresolution}. 
\dalle is the latest VLM developed by OpenAI, leveraging advanced transformer architectures to generate images from textual descriptions.
\sd, developed by StablilityAI, is a text-to-image generation VLM that uses diffusion processes to create images from textual descriptions. 

To ensure consistency and equity in using VLMs, we design the same prompt template to prompt the selected VLMs. 
Table~\ref{tab:promptPhrase} lists the main prompt phrases and relevant examples, which involve common laptop models, the number, size, style, and location of generated stickers, and other detailed image content information. The prompt template consists of these phrases, e.g., \textit{``a laptop, three stickers on the lid, one small music sticker, one small animal sticker, one small superhero sticker, stickers close to each other, top-down view, screen turn off, on a flat surface, a solid dark background"}. 
Note that to enhance the generality of the synthesized dataset, the prompt template omits information about the quality of the generated images, e.g., \textit{sharp focus} or \textit{highly detailed}, thereby generating images with varying levels of clarity that can be used to test and evaluate the generalization capability of \sdm{s}.

\subsection{Adversarial Image Generation}\label{subsec:adv}
DNNs are vulnerable to adversarial attacks, which are imperceptible to humans, but easily misclassified by DNNs~\cite{madry2017towards}. To study the robustness of \sdm{s} against adversarial attacks and how these attacks affect the prediction uncertainty of \sdm{s}, we construct an adversarial image dataset by perturbing images in an original dataset. 
Specifically, given an image $X$ containing $N$ targets $\mathcal{T}=\{t_1, t_2, ..., t_{N}\}$, each target is labeled by a ground-truth $l_n \in \{sticker, logo, background\}$, an adversarial attacker aims to inject perturbations on $X$ to obtain a perturbed image $X^\prime$ that satisfies:
\begin{equation}
    \forall 1 \leq n \leq N \wedge t_n \in \mathcal{T}: f(X^\prime, t_n)=l_n^\prime \neq f(X, t_n)=l_n,
\end{equation}
where $f$ denotes the \sdm. As a result, the generated adversarial image $X^\prime$ makes all targets incorrectly predicted. We adopt an adversarial attack approach called Dense Adversary Generation (DAG)~\cite{xie2017adversarial} to construct the adversarial image dataset. DAG aims to generate adversarial examples for semantic segmentation and object detection tasks where multiple objects must be recognized. It requires an original dataset to perturb the images in it. We use the dataset to evaluate the pre-trained \sdm as the original dataset and then add adversarial perturbations to the images in the original dataset to obtain an adversarial image dataset. \looseness=-1

\begin{table}[!t]
    \centering
    \caption{Prompt Phrase with Example}
    \resizebox{0.6\linewidth}{!}{\begin{tabular}{ll}
\toprule
\textbf{Phrase}  & \textbf{Example}                                                                  \\
\midrule
laptop model     & \textit{"a laptop", "a MacBook laptop", "a Dell laptop", "an HP laptop", ...}     \\
sticker number   & \textit{"one", "two", "three", ...}                                               \\
sticker size     & \textit{"small", "medium", "big"}                                                 \\
sticker style    & \textit{"Apple logo", "Dell logo", ..., "cartoon character", "animal", ...}       \\
sticker location & \textit{"close to the center of the laptop", "stickers close to each other", ...} \\
other            & \textit{"top-down view", "screen turn off", "a solid dark background", ...}       \\
\bottomrule
\end{tabular}
}
    \label{tab:promptPhrase}
\end{table}

\section{Experiment Design} \label{subsec:exp_design}
\subsection{Sticker Detection Models}
We obtain six pre-trained \sdm{s} employed in the robotic software, representing state-of-the-art DNN-based object detection models. These models are designed by considering multiple model architectures which are \fasterrcnn~\cite{ren2015faster}, \fasterrcnnv~\cite{li2021benchmarking}, \retinanet~\cite{lin2017focal}, \retinanetv~\cite{zhang2020bridging}, \ssd~\cite{liu2016ssd}, and \ssdlite~\cite{howard2019searching}. DTI trained these models using a specially designed sticker detection dataset containing thousands of labeled laptop images with stickers in various poses, sizes, and lighting conditions. All six \sdm{s} are trained based on the open-source implementation in PyTorch~\cite{paszke2019pytorch}.

We modify the six pre-trained models to be compatible with the MC-Dropout method by injecting prediction-time activated dropout layers. Specifically,  for \fasterrcnn, \fasterrcnnv, \retinanet, and \retinanetv, we add dropout layers to the convolutional layers of the Feature Pyramid Network, which is a common component of these four models and built on the backbones of these four models. Regarding \ssd and \ssdlite, we add dropout layers to their detection heads. In our experiments, we choose 9 dropout ratios from 0.1 to 0.5 with an interval of 0.05 to study the effect of dropout rates on the performance of \sdm{s}.

\subsection{Benchmark Datasets}
We obtain a dataset containing 150 images from DTI, i.e., \origimg, and synthesize two datasets by prompting two VLMs, \dalle and \sd, i.e., \dalleimg and \stableimg, each containing 150 images. By applying the adversarial attack method (DAG) to \origimg, \dalleimg, and \stableimg, we generate three adversarial datasets: \advorigimg, \advdalleimg, and \advstableimg. Specifically, for each image $X$ in each dataset, we execute DAG 10 times and obtain 10 adversarial images for $X$, and therefore, each adversarial dataset contains 1500 images. These three adversarial datasets are used to assess the robustness of the \sdm{s}.

\subsection{Evaluation Metrics}
\textbf{\textit{Mean Average Precision (\map)}} is a metric for evaluating the accuracy of object detectors~\cite{everingham2010pascal} and is calculated as:
\vspace{-2pt}
\begin{equation}
\vspace{-2pt}
    \map=\frac{1}{nc}\sum_{c=1}^{nc}AP_c,
\end{equation}
where $AP_c$ is the average precision for class $c$ and $nc$ is the number of classes. $AP$ is the area under the precision-recall curve $p(r)$ that is calculated as $\int_{0}^{1} p(r) dr$.
For object detection tasks, precision measures how accurate the predictions are and is calculated as the number of true positives divided by the number of all detected objects, while recall measures how good the model is at recalling classes and is defined as the number of true positives divided by the number of all ground-truth objects. True positives are determined based on Intersection over Union (IoU), which quantifies how close the predicted and ground-truth bounding boxes are by taking the ratio between the area of intersection and the area of the union of the predicted and ground-truth boxes: $IoU=(A_{P}\cap A_{G})/(A_{P}\cup A_{G})$, where $A_{P}$ and $A_{G}$ are the area of the predicted and ground-truth boxes, respectively. Then, if the IoU of the predicted and true boxes is greater than a threshold and the object is correctly classified, they are considered a match and therefore a true positive. We set the threshold to 0.5, commonly used for object detection~\cite{catak2021prediction}. \looseness=-1

\noindent \textbf{\textit{UQ Metrics (\uqms{s})}} are the UQ metrics defined in Section~\ref{subsec:uqmetrics}, which includes \vr, \se, \mi, \tv, and \ps.

\noindent \textbf{\textit{Robustness Score (\rs)}} measures the robustness of each \sdm in performing the sticker detection task. 
We calculate \rs regarding \map and \uqms, respectively, i.e., \rsmap and \rsuqm by considering the model's performance on dataset $\mathcal{D}$ and $\mathcal{D}$'s adversarial version $\mathcal{D}_{adv}$. 
Specifically, for image $X \in \mathcal{D}$, there are 10 adversarial images $\{X_{adv}^1, X_{adv}^2, ..., X_{adv}^{10}\}$ from $\mathcal{D}_{adv}$, and we consider \rs regarding metric $\mathcal{M}$ to be high when the model performs well in terms of $\mathcal{M}$ (i.e., high \map or low \uqms) and the difference in $\mathcal{M}$ between $X$ and $X$'s 10 adversarial examples is small. We then calculate \rsmap as:
\vspace{-2pt}
\begin{equation}
\vspace{-2pt}
    RS_{mAP}=Avg_{mAP}-Diff_{mAP},
\end{equation}
where $Avg_{mAP}$ is the mean value of \map calculated by averaging the value achieved on $X$ and $X$'s 10 adversarial examples: $(mAP_X+\sum_{i=1}^{10}mAP_{X_{adv}^i})/11$; $Diff_{mAP}$ measures how differently the model performs on $X$ and $X$'s 10 adversarial examples in terms of \map: $\sum_{i=1}^{10}|mAP_X-mAP_{X_{adv}^i}|/10$. 
For \rsuqm, we first calculate \rs for each uncertainty metric $\mathcal{M}$:
\vspace{-8pt}
\begin{equation}
\vspace{-2pt}
    RS_{\mathcal{M}} = 1 - (Avg_\mathcal{M}+Diff_\mathcal{M}),
\end{equation}
where $\mathcal{M}\in\{\vr, \se, \mi, \tv, \ps\}$. $Avg_\mathcal{M}$ and $Diff_\mathcal{M}$ are calculated using the same equations as $Avg_{mAP}$ and $Diff_{mAP}$. We then calculate \rsuqm by taking the mean value of \rs for each uncertainty metric:
\vspace{-3pt}
\begin{equation}
\vspace{-1pt}
    RS_{uq} = \frac{\sum_{\mathcal{M} \in UQMs}RS_{\mathcal{M}}}{count(UQMs)}.
\end{equation}
\rsuqm combines the \rs of the five \uqms{s} to show the overall robustness of the \sdm{s} in terms of uncertainty.

\begin{table}[!ht]
    \centering
    \caption{Datasets, metrics, and statistical tests for each RQ}
    \resizebox{0.6\columnwidth}{!}{\begin{tabular}{llll}
\toprule
\textbf{RQ} & \textbf{Dataset} & \textbf{Metric} & \textbf{Statistical Test} \\ \midrule
RQ1 & \begin{tabular}[c]{@{}l@{}}\origimg, \dalleimg, \stableimg \end{tabular} & \map        & \multirow{6}{*}{\begin{tabular}[c]{@{}l@{}}Friedman test, \\Wilcoxon Signed-Rank test, \\rank-biserial correlation, \\Holm–Bonferroni method\end{tabular}} \\ \cmidrule(r){1-3}
RQ2 & \begin{tabular}[c]{@{}l@{}}\origimg, \dalleimg, \stableimg \end{tabular}     & \begin{tabular}[c]{@{}l@{}}\vr, \se, \mi,\\ \tv, \ps\end{tabular}        &                  \\ \cmidrule(r){1-3}
RQ3 & \begin{tabular}[c]{@{}l@{}}\origimg, \advorigimg, \dalleimg, \\\advdalleimg, \stableimg, \advstableimg \end{tabular}     & \begin{tabular}[c]{@{}l@{}}\rsmap, \rsuqm\end{tabular}          &                  \\ \midrule
RQ4 & \begin{tabular}[c]{@{}l@{}}\origimg, \dalleimg, \stableimg \end{tabular}     & \begin{tabular}[c]{@{}l@{}}\map, \vr, \se,\\ \mi, \tv, \ps\end{tabular}          &  \begin{tabular}[c]{@{}l@{}}Spearman's rank correlation, \\Holm–Bonferroni method\end{tabular}    \\ \bottomrule
\end{tabular}}
    \label{tab:exp_design}
\end{table}

\subsection{Research Questions}
We answer the following four research questions (RQs):
\textbf{\textit{RQ1:}} How does each \sdm perform when detecting stickers regarding detection accuracy? 
\textbf{\textit{RQ2:}} How does each \sdm perform when detecting stickers regarding prediction uncertainty?
\textbf{\textit{RQ3:}} How robust is each \sdm in detecting stickers?
\textbf{\textit{RQ4:}} How does the detection accuracy correlate with the prediction uncertainty?
Table~\ref{tab:exp_design} describes the employed datasets, metrics, and statistical tests for answering RQs.

\vspace{-6pt}

\subsection{Statistical Test}
\vspace{-2pt}
Since we compare more than two paired groups, i.e., \sdm{s} using the same image dataset, we conduct the statistical analysis using the Friedman test~\cite{friedman1937use}, the Wilcoxon Signed-Rank test~\cite{wilcoxon1992individual}, and rank-biserial correlation~\cite{kerby2014simple}. Following the guideline~\cite{mangiafico2016summary,garcia2010advanced}, we employ the Friedman test to verify if there are overall significant differences among all the paired groups (i.e., the models). If at least one group differs significantly from the others, $p < 0.05$ will be computed by this statistical test. If this is the case, we apply the Wilcoxon Signed-Rank test for pairwise comparisons (i.e., comparing two models) with a significance level of 5\%. To interpret the Wilcoxon Signed-Rank test more precisely, we calculate the median and the standard deviation by the median absolute deviation (MAD) of our data, i.e., [median $\pm$ 2 MAD]~\cite{tukey1977exploratory,reimann2005background}. It checks if the underlying distribution is stochastically equal to, less than, or greater than a distribution symmetric about zero, and accordingly defines the relevant alternative hypothesis for more effective execution of the Wilcoxon Signed-Rank test. Moreover, for pairwise groups with significant differences, we use the rank-biserial correlation as the effect size of the Wilcoxon Signed-Rank test to determine which group shows better results. The rank-biserial correlation ranges from -1 to 1, where a positive result indicates that the first group tends to be larger than the second, and a negative result means the opposite, whereas 0 explains there is no significant difference between the two paired groups. 

To study the correlation between \map and \uqms{s} for RQ4, we use Spearman's rank correlation coefficient~\cite{spearman1961proof}, which provides the correlation coefficient ($\rho$) and the significance level ($p$). A $p$ less than 0.05 indicates a significant correlation between \map and \uqms{s}, whereas $\rho$ ranges from -1 to 1, revealing the direction and strength of the correlation. A positive $\rho$ indicates a positive correlation, meaning that one variable increases as the other increases, and vice versa. A $\rho$ of 0 shows no correlation. 
Considering that the Wilcoxon Signed-Rank test and Spearman's rank correlation coefficient are utilized multiple times to compare multiple paired groups, we address the issue of aggregated error probabilities resulting from multiple comparisons by applying the Holm–Bonferroni method~\cite{holm1979simple}. This multiple comparison correction technique adjusts the significance level from the overall and controls the family-wise error rate at an $\alpha$ level, e.g., 5\%. 

\section{Results and Analyses}\label{sec:results}
We present descriptive statistics and summarize the key results of the statistical tests in Sections~\ref{subsec:rq1} to~\ref{subsec:rq4}, and then provide guidelines for selecting \sdm{s} in Section~\ref{subsec:guidelines}. 
The complete results are available in our replication package~\cite{github}.

\subsection{Results for RQ1 -- Mean Average Precision}\label{subsec:rq1}
\begin{figure}[!t]
\centering
\includegraphics[width=0.6\linewidth]{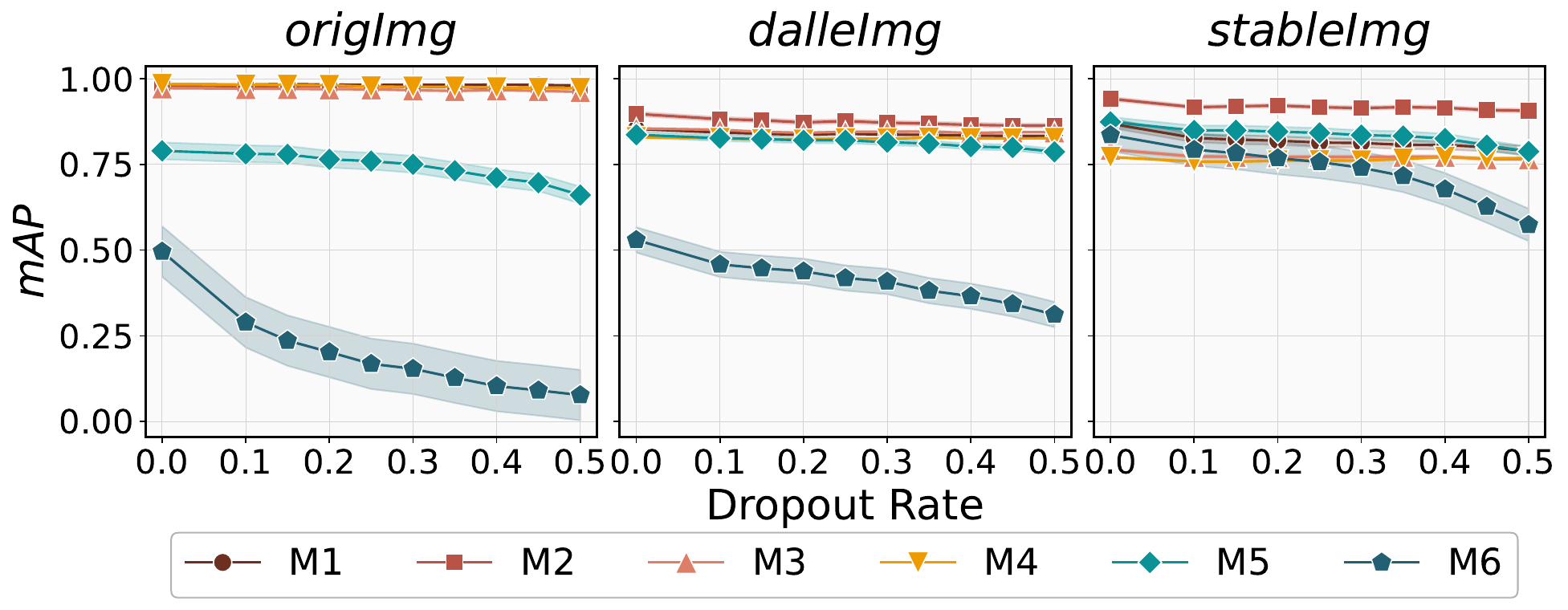}
\caption{Average \map achieved by \sdm{s} for \origimg, \dalleimg, and \stableimg at different dropout rates. M1: \fasterrcnn, M2: \fasterrcnnv, M3: \retinanet, M4: \retinanetv, M5: \ssd, and M6: \ssdlite \xspace -- \textit{RQ1}.}
\label{fig:mAP}
\end{figure}

Figure~\ref{fig:mAP} shows the \map results achieved by each \sdm for each dataset at different dropout rates with 95\% confidence intervals. For the \origimg dataset, \fasterrcnn, \fasterrcnnv, \retinanet, and \retinanetv all achieved \map values close to 1 across all dropout rates, which indicates these four \sdm{s} achieve high precision and accuracy in detecting stickers correctly, and their performance is stable as the dropout rate increases. Lower \map values are observed for \ssd and \ssdlite, with \ssdlite consistently performing the worst. Moreover, the \map values for both \sdm{s} (\ssd and \ssdlite) decrease as the dropout rate increases, indicating that they are more susceptible to the dropout rate. For \dalleimg dataset, \fasterrcnnv achieved the consistent best performance at all dropout rates. \fasterrcnn, \retinanet, and \retinanetv performed comparably, which slightly outperformed \ssd. \ssdlite performed consistently the worst. For \stableimg dataset, we can observe that at all dropout rates, \fasterrcnnv performed the best followed by \ssd and \fasterrcnn. \ssdlite outperformed \retinanet and \retinanetv at dropout rates below 0.2, after which the performance of \ssdlite started to decrease significantly and underperformed \retinanet and \retinanetv, while \retinanet and \retinanetv performed consistently across all dropout rates. 
Besides, when comparing the \map achieved by \sdm{s} on LLM-synthetic datasets (\dalleimg and \stableimg) and \origimg, we observe that \fasterrcnn, \fasterrcnnv, \retinanet, and \retinanetv perform poorer on the LLM-synthetic datasets than on \origimg dataset, while \ssd and \ssdlite exhibit better performance on the LLM-synthetic datasets than on \origimg dataset.

We rank the best \sdm that significantly outperformed the other \sdm{s} on each metric according to the pairwise comparison results. For \sdm{s} having no significant differences, we consider them to be a tie. Table~\ref{tab:bestModel_summary} shows the results of best \sdm{(s)} for each metric, where the last row shows the \sdm recommendations based on different metrics. 
The best \sdm for \map is shown in the second column of Table~\ref{tab:bestModel_summary}. 
We can observe that for \origimg dataset, \fasterrcnn, \fasterrcnnv, and \retinanetv all performed the best in terms of \map, significantly outperforming the other \sdm{s} (\retinanet, \ssd, and \ssdlite) across all the dropout rates. As for \dalleimg and \stableimg datasets, \fasterrcnnv significantly outperforms the other five \sdm{s}. When looking at the best \sdm across all datasets, \fasterrcnnv is recommended for all three datasets.

\conclusion{
\textbf{Conclusion for RQ1}: Different \sdm{s} perform differently in terms of \map on different datasets. Specifically, \fasterrcnn, \fasterrcnnv, and \retinanetv all perform the best for \origimg dataset, while \fasterrcnnv achieves the best performance for \dalleimg and \stableimg datasets. Regarding the best \sdm across all datasets, \fasterrcnnv achieves the overall best performance regarding \map and is recommended.
}

\begin{table}[!ht]
\caption{Best \sdm{(s)} across different metrics and datasets for all dropout rates based on pairwise comparison results. M1: \fasterrcnn, M2: \fasterrcnnv, M3: \retinanet, M4: \retinanetv, M5: \ssd, and M6: \ssdlite. 
M$i$-$j$ means that M$i$, M$i+1$, ..., M$j$ all performed the best and had no significant differences \xspace -- \textit{RQ1}, \textit{RQ2}, and \textit{RQ3}.
}
\label{tab:bestModel_summary}
\centering
\resizebox{0.6\linewidth}{!}{
\begin{tabular}{lllllllll}
\toprule
\multirow{2}{*}{\textbf{Dataset}} & \textbf{RQ1} & \multicolumn{5}{l}{\textbf{RQ2}} & \multicolumn{2}{l}{\textbf{RQ3}} \\ \cmidrule(r){2-2} \cmidrule(r){3-7} \cmidrule(r){8-9}
 & \textbf{\map} & \textbf{\vr} & \textbf{\se} & \textbf{\mi} & \textbf{\tv} & \textbf{\ps} & $\bm{\mathit{RS_{mAP}}}$ & $\bm{\mathit{RS_{uq}}}$ \\ \midrule
\origimg & M1,2,4 & M1,2,5,6 & M1,4 & M4 & M4 & M4 & M1,4 & M4 \\
\dalleimg & M2 & M1-6 & M4 & M4 & M4 & M4 & M2 & M4 \\
\stableimg & M2 & M1-6 & M4 & M4 & M4 & M4 & M2 & M4 \\ \midrule
$\bm{\mathit{Rec_{metric}}}$ & M2 & M1,2,5,6 & M4 & M4 & M4 & M4 & M2 & M4 \\ \bottomrule
\end{tabular}
}
\end{table}

\subsection{Results for RQ2 -- Uncertainty}\label{subsec:rq2}
\begin{figure}[!ht]
\centering
\includegraphics[width=0.6\linewidth]{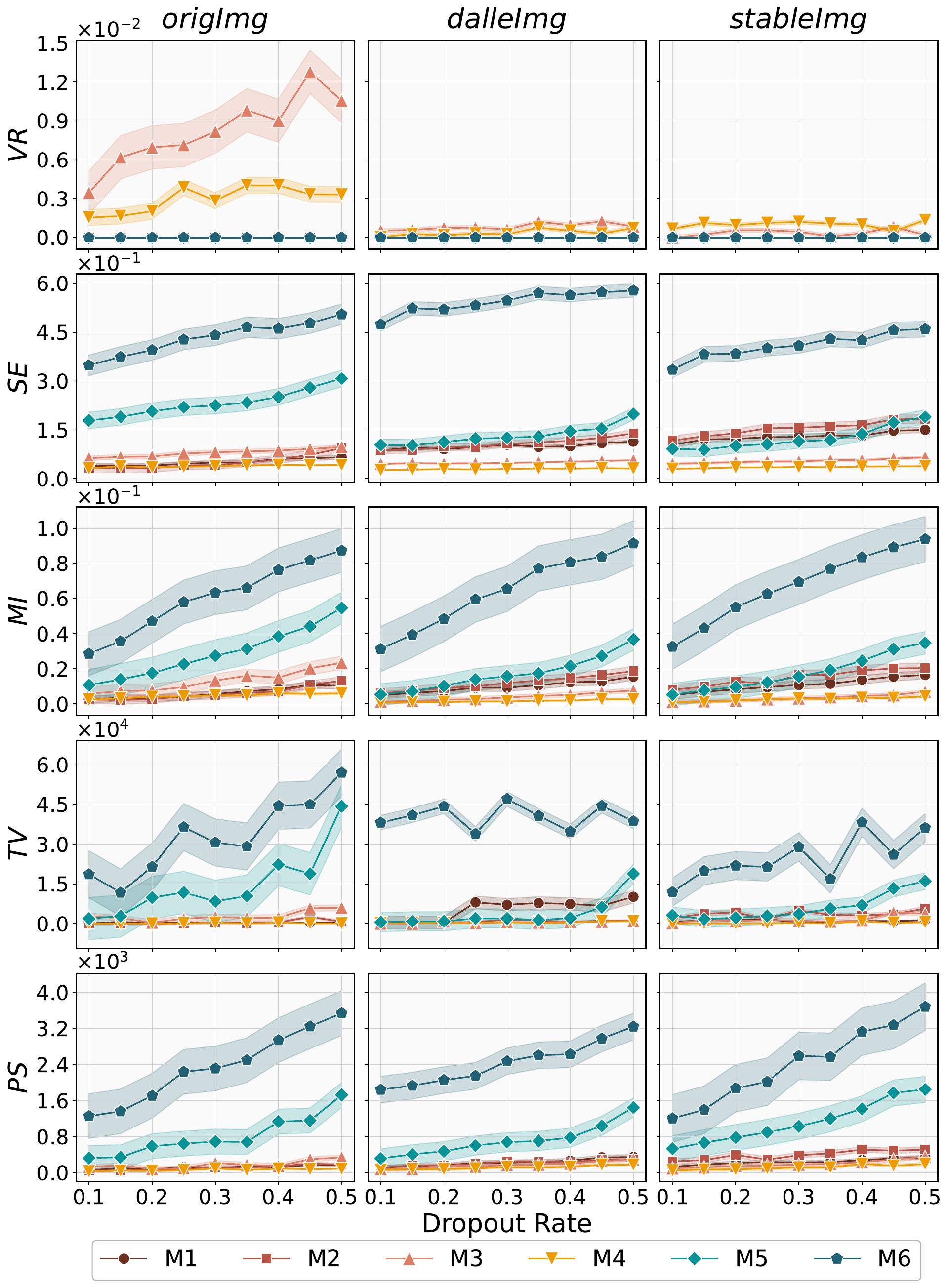}
\caption{Average \uqms{s} achieved by \sdm{s} for \origimg, \dalleimg, and \stableimg at different dropout rates. M1: \fasterrcnn, M2: \fasterrcnnv, M3: \retinanet, M4: \retinanetv, M5: \ssd, and M6: \ssdlite \xspace -- \textit{RQ2}.}
\label{fig:UQMs}
\end{figure}

Figure~\ref{fig:UQMs} presents the mean results of \uqms{s} for each dataset at different dropout rates with 95\% confidence intervals. 

\textit{Uncertainty for Label Classification.} We have the following observations regarding three label classification \uqms{s} (i.e., \vr, \se, and \mi). For \vr, \fasterrcnn, \fasterrcnnv, \retinanet, \retinanetv, \ssd, and \ssdlite all achieve a very low \vr, i.e., less than 0.015 for all three datasets at all dropout rates. The results suggest that all \sdm{s} exhibit negligible uncertainty in label classification dispersion in the three datasets, that is, when an object is detected multiple times, the \sdm{s} are confident that they will give the same classification result. However, for \origimg dataset, we observe slightly higher \vr values, which increase as the dropout rate increases for both \retinanet and \retinanetv, showing higher uncertainty and less stability of these two \sdm{s} in terms of \vr as dropout rate increases. As for \se and \mi, \ssdlite consistently achieves the highest uncertainty values of both \uqms{s} for all datasets at all dropout rates, followed by \ssd, which exhibits high uncertainty values regarding these two \uqms{s} for \origimg and \dalleimg datasets. However, exceptions can be observed for \stableimg dataset, where \ssd shows lower \se and \mi values than \fasterrcnn and \fasterrcnnv at low dropout rates and starts to show higher uncertainty as the dropout rate increases. Besides, \fasterrcnn, \fasterrcnnv, \retinanet, and \retinanetv achieve comparable performance in terms of \se and \mi for \origimg dataset at all dropout rates, while when it comes to \dalleimg and \stableimg datasets, \retinanet and \retinanetv consistently achieve the lowest \se and \mi values.
Recall that \se quantifies the amount of information in the predictions, while \mi measures the difference in information between the prediction and the posterior. The results for \se and \mi show that the prediction of \ssd and \ssdlite contain more information and differ more from the posterior, thus exhibiting higher uncertainty in label classification as compared to the other four \sdm{s}, i.e., \fasterrcnn, \fasterrcnnv, \retinanet, and \retinanetv.

\textit{Uncertainty for Bounding Box Regression.} Similar observations can be made for the bounding box regression \uqms{s}, that is, \ssd and \ssdlite show higher \tv and \ps values for all datasets at all dropout rates except for \tv on \dalleimg dataset, where \ssd achieves lower \tv than \fasterrcnn until the dropout rate reaches 0.45. Besides, \fasterrcnn, \fasterrcnnv, \retinanet, and \retinanetv achieve comparable and consistent low \tv and \ps, except for \tv of \fasterrcnn on \dalleimg dataset. The results of \tv and \ps show that \ssd and \ssdlite exhibit higher uncertainty in predicting bounding boxes than the other four \sdm{s}. Moreover, when the dropout rate increases, the uncertainties in \ssd and \ssdlite also increase, while for the other \sdm{s}, the dropout rate has little effect on their uncertainties.

Table~\ref{tab:bestModel_summary} shows the best \sdm{(s)}, where we can find that regarding \vr, \fasterrcnn, \fasterrcnnv, \ssd, and \ssdlite achieve the best performance on \origimg dataset, while all \sdm{s} perform comparable on \dalleimg and \stableimg datasets. For \se, \mi, \tv, and \ps, \retinanetv perform the best, i.e., shows the lowest uncertainty, except for \se on \origimg dataset, where \fasterrcnn and \retinanetv achieve comparable best performance. 
When looking at the \sdm recommendations across all datasets, regarding \vr, \fasterrcnn, \fasterrcnnv, \ssd, and \ssdlite are recommended, while for \se, \mi, \tv, and \ps, \retinanetv is recommended as the best \sdm. 

\conclusion{
\textbf{Conclusion for RQ2}: Different \sdm{s} show different levels of uncertainties for different \uqms{s}. Regarding \vr, \fasterrcnn, \fasterrcnnv, \ssd, and \ssdlite are recommended as the best \sdm{s} since they have the lowest uncertainties across all datasets, while for \se, \mi, \tv, and \ps, \retinanetv achieves the lowest uncertainties and is recommended as the best \sdm.
}

\begin{figure}[!ht]
\centering
\includegraphics[width=0.6\linewidth]{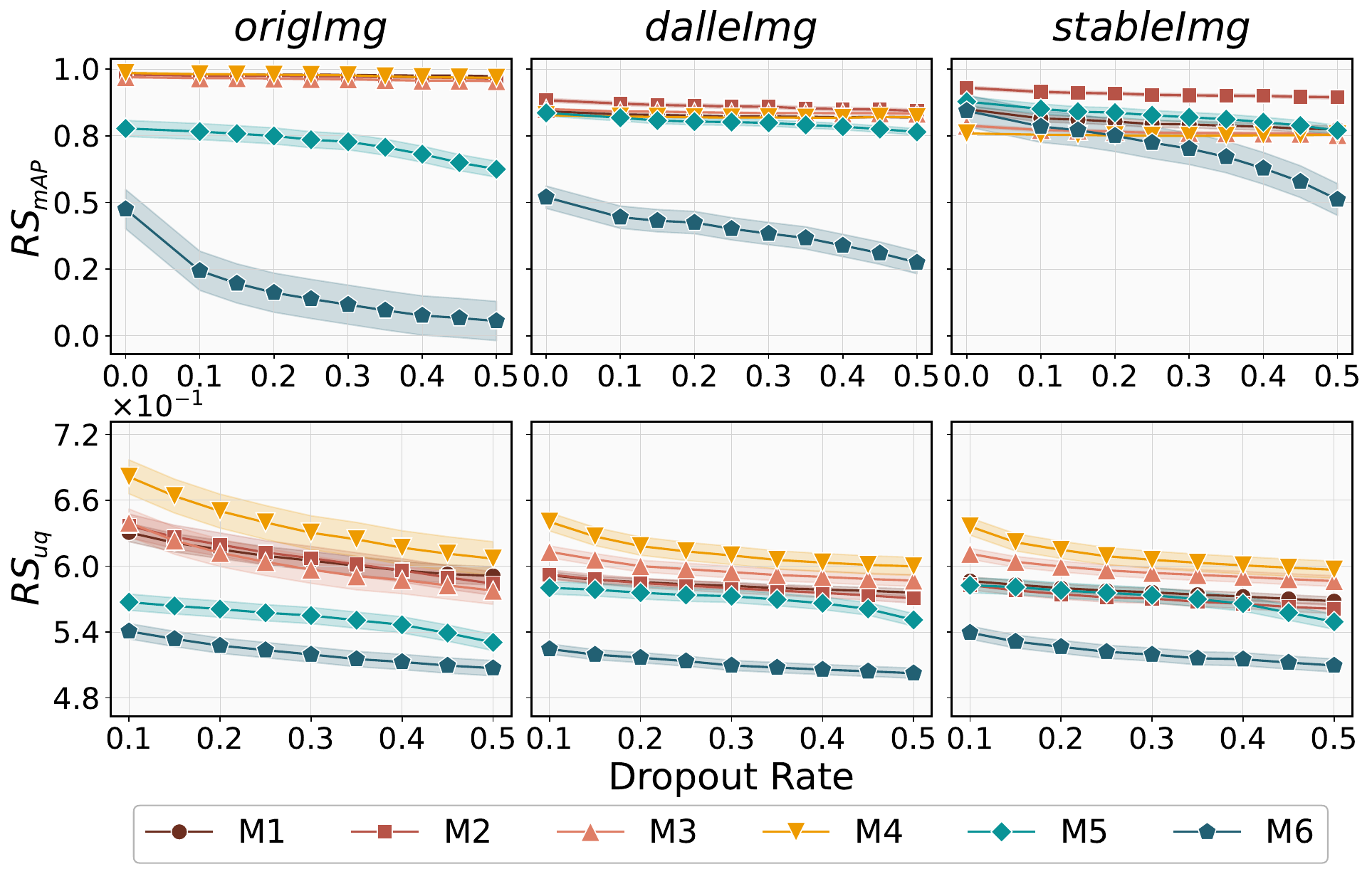}
\caption{Average \rsmap and \rsuqm achieved by \sdm{s} for \origimg, \dalleimg, and \stableimg at different dropout rates. M1: \fasterrcnn, M2: \fasterrcnnv, M3: \retinanet, M4: \retinanetv, M5: \ssd, and M6: \ssdlite \xspace -- \textit{RQ3}.}
\label{fig:RS}
\end{figure}

\subsection{Results for RQ3 -- Robustness}\label{subsec:rq3}
Figure~\ref{fig:RS} shows the mean \rsmap and \rsuqm results with 95\% confidence intervals. Regarding \rsmap, we observe that \fasterrcnnv performs the best for \dalleimg and \stableimg datasets at all dropout rates, while for \origimg dataset, \fasterrcnn, \fasterrcnnv, \retinanet, and \retinanetv all perform the best with a \rsmap close to 1. \ssdlite consistently performs the worst for \origimg and \dalleimg datasets at all dropout rates, and performs worst on \stableimg when the dropout rate is above 0.2. Besides, the increase in dropout rate has different effects on different \sdm{s}, for example, the dropout rate increase leads to a decrease of \rsmap for \ssdlite, while does not have much impact on \fasterrcnn, \fasterrcnnv, \retinanet, and \retinanetv.
Regarding \rsuqm, we observe that the results are consistent for all three datasets at all dropout rates, that is, \retinanetv achieves the best \rsuqm, while \ssdlite performs the worst. We also see a decrease in \rsuqm as the dropout rate increases. We also show the best \sdm{(s)} in Table~\ref{tab:bestModel_summary}. For \rsmap, \fasterrcnn and \retinanetv perform the best on \origimg dataset, while \fasterrcnnv perform the best on \dalleimg and \stableimg datasets. When looking at the best \sdm across all datasets, \fasterrcnnv is recommended. Regarding \rsuqm, \retinanetv performs consistently the best across all three datasets and is therefore recommended. 

\conclusion{
\textbf{Conclusion for RQ3}: The six \sdm{s} perform differently for different datasets in terms of \rsmap and \rsuqm. Specifically, regarding \rsmap, \fasterrcnn and \retinanet achieve the best performance on \origimg dataset, and \fasterrcnnv perform the best on \dalleimg and \stableimg datasets. Regarding \rsuqm, \retinanetv performs consistently the best across all three datasets.
}

\vspace{1pt}
\subsection{Results for RQ4 -- Correlation}\label{subsec:rq4}
Table~\ref{tab:spearmanModel} shows the results of Spearman's rank correlation test for the correlation between \map and the five \uqms{s} for different \sdm{s}. As for the correlation between \map and \vr, for all \sdm{s} the difference is either insignificant or non-existent, while in terms of the correlation between \map and the other four \uqms{s} (\se, \mi, \tv, and \ps), negative correlations can be observed for all \sdm{s}, except for \ssd, which achieves insignificant correlation between \map and \tv/\ps. This observation suggests that as uncertainty increases (except for \vr), detection precision and accuracy will decrease. In addition, we also observe that as the dropout rate increases, the correlation between \map and \se, \mi, \tv, and \ps increases. This is reasonable since higher dropout rates will introduce higher uncertainties in the predictions.

\conclusion{
\textbf{Conclusion for RQ4}: We observe significant negative correlations between \map and \se, \mi, \tv, and \ps for all \sdm{s}, except for \tv and \ps achieved by \ssd, while insignificant or non-existent correlations are observed for the correlation between \map and \vr. Besides, the correlation between \map and \se, \mi, \tv, and \ps increases as the dropout rate increases.
}

\begin{table}[!ht]
\caption{Results of the Spearman's rank correlation coefficient between \map and \uqms{s} for each \sdm{} on all dropout rates and datasets. A \underline{value} or \textit{N/A} denotes that the correlation is not statistically significant or non-existent, 
otherwise, it is significant, i.e., $p < 0.05$ \xspace -- \textit{RQ4}.}
\label{tab:spearmanModel}
\centering
\resizebox{0.6\linewidth}{!}{
\begin{tabular}{lrrrrr}
\toprule
\textbf{\sdm{}} & \textbf{\vr} & \textbf{\se} & \textbf{\mi} & \textbf{\tv} & \textbf{\ps} \\ \midrule
\fasterrcnn & N/A & -0.312 & -0.277 & -0.476 & -0.461 \\
\fasterrcnnv & N/A & -0.408 & -0.357 & -0.365 & -0.334 \\
\retinanet & \underline{0.012} & -0.161 & -0.107 & -0.328 & -0.329 \\
\retinanetv & \underline{0.098} & -0.209 & -0.065 & -0.316 & -0.295 \\
\ssd & N/A & -0.530 & -0.455 & \underline{-0.016} & \underline{0.047} \\
\ssdlite & N/A & -0.510 & -0.070 & -0.269 & -0.174 \\ \bottomrule
\end{tabular}
}
\end{table}

\subsection{Concluding Remarks and Guidelines}\label{subsec:guidelines}
\begin{figure}[!ht]
\centering
\includegraphics[width=0.6\linewidth]{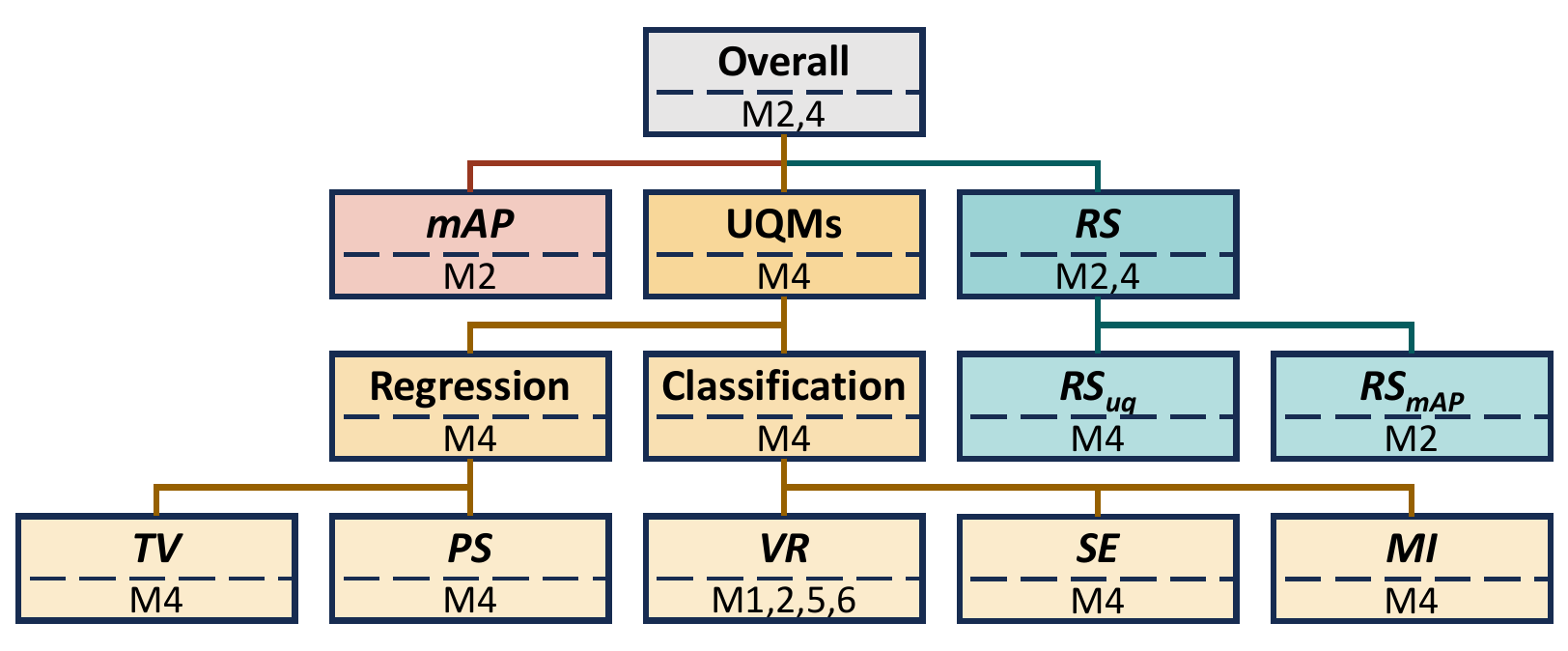}
\caption{Guidelines for Selecting Sticker Detection Models.}
\label{fig:guideline}
\end{figure}

We analyze the performance of each \sdm regarding detection accuracy (\map), prediction uncertainty (\uqms), and adversarial robustness (\rsmap and \rsuqm). Based on the results, we observe that different \sdm{s} perform differently regarding different metrics. In practice, considering different purposes and applications, selecting the appropriate \sdm based on specific performance requirements is crucial. Therefore, we provide guidelines for selecting \sdm{s} from various perspectives in Figure~\ref{fig:guideline}. Regarding detection accuracy, i.e., \map, \fasterrcnnv is recommended as it achieves the overall best performance. As for \uqms{s}, for both label classification and bounding box regression tasks, we recommend \retinanetv as it shows lower uncertainty in terms of both types of uncertainty metrics. For the robustness of \sdm{s}, \fasterrcnnv shows the overall highest robustness regarding detection accuracy, i.e., \rsmap, and is recommended as the best \sdm. When considering robustness regarding uncertainty, \retinanetv is recommended as the best \sdm. Based on the guidelines, adaptive software integrated within the robot to select \sdm{s} would significantly enhance the robot's decision-making capabilities, particularly in dynamic and uncertain environments. Such software can automatically select the most appropriate \sdm{s} by considering various metrics.

\section{Threats to Validity}
\textit{Conclusion Validity} concerns the reliability of the conclusions. We employed appropriate statistical tests to draw reliable conclusions and followed a rigorous statistical procedure to analyze the collected data.
\textit{Internal Validity} is related to the parameter settings. To determine the hyperparameters for the clustering algorithm, i.e., HDBSCAN, we ran different combinations of hyperparameter values on a small subset of the original dataset and selected the combination that correctly clustered all objects. Besides, the threshold value for IoU is set to 0.5, following Catak et al.~\cite{catak2021prediction}. 
\textit{Constructive Validity} concerns the metrics used for the evaluation and we employed comparable metrics to ensure fair comparisons. Specifically, we used \map to measure the sticker detection performance and five \uqms{s} commonly used in classification and regression to measure model uncertainty. We also calculated \rs to measure the robustness of the models under adversarial attacks.
\textit{External Validility} is about the generalizability of the empirical evaluation. We employ six models and six datasets. These models are based on different architectural designs and represent state-of-the-art object detection models. Regarding the benchmark datasets, in addition to the original dataset, we synthesize two datasets by prompting two VLMs, i.e., \dalle and \sd. Besides, we generate adversarial datasets by perturbing the original and synthetic datasets.

\section{Lessons Learned and Industrial Perspectives}

\textit{Understanding, Utilizing, and Handling Uncertainty in Refurbishing Laptop software.}
UQ plays a pivotal role in enhancing the refurbishment of laptops by guiding laptop image data collection, ensuring dataset diversity, and determining the need for manual intervention. 
Collecting laptop images is time-consuming and significantly impacts solution quality, and UQ can guide decisions on whether additional data is needed. Specifically, UQ helps identify areas with high prediction uncertainty, indicating the need for more diverse and comprehensive training data. 
Besides, UQ is essential in addressing the limitations of existing datasets created by DTI, which primarily features Lenovo laptops and common sticker placements, e.g., stickers on the back cover. Also, images in the DTI dataset were captured in a relatively sterile setup with high quality that may not reflect real-world conditions. Thus, \sdm{s} need to be tested under different conditions in handling unexpected situations, such as stickers near a mouse pad, where processor branding stickers are often found. By highlighting gaps in data diversity, UQ directs efforts towards collecting images from various laptop brands and unexpected sticker locations, ensuring \sdm{s} are robust and adaptable to real-world conditions. Finally, UQ informs the need for manual intervention in the refurbishment process, enabling efficient and reliable operations by indicating when human oversight is required.
In addition to the uncertainty in \sdm{s}, the overall process of laptop refurbishing experiences uncertainties from various aspects, such as humans working in collaboration with robots, hardware errors, and other environmental factors. 
Hence, a holistic UQ method is required for the entire laptop refurbishing process that quantifies uncertainties from various aspects and eventually provides an overall uncertainty. 
Such quantified uncertainty plays a key role in identifying factors contributing to uncertainties, followed by devising best practices to reduce the overall uncertainty.

\textit{Realism of VLM-Generated Images.}
To assess the \sdm{s}, we generated images of laptops with stickers using two commonly used pre-trained VLMs. However, this poses an open question of whether the images generated are realistic. In our context, all the generated images were manually checked for realism. However, such an approach is not scalable, especially when generating a large-scale dataset. To this end, we foresee the need for an automated method to generate realistic images of laptops with stickers. Pre-trained VLMs already possess a strong understanding of general image features, and domain-specific fine-tuning can further refine their output. Thus, one possible solution is to fine-tune VLMs to adapt to specific domains related to laptops and stickers, thereby helping produce more realistic images from the outset and reducing the need for extensive post-production filtering. Besides, large language models can be another way of realism evaluation~\cite{wu2024reality}. Furthermore, training a realism classifier to distinguish realistic from unrealistic images can automate the manual check process. In the future, we plan to investigate scalable and efficient methods to generate large-scale realistic laptop image datasets.

\section{Related Works}


\textit{UQ in DNNs.} UQ helps enhance model reliability and trustworthiness. Various UQ methods have been studied to measure confidence in the model predictions. BNN provides a probabilistic framework for DNNs through Bayesian inference~\cite{tran2019bayesian}. It is systematic but computationally expensive and has a complex implementation. As a Bayesian approximation, MC-Dropout is a practical and widely used UQ method~\cite{gal2016dropout}, which reduces the computational cost. Deep Ensembles (DE)~\cite{lakshminarayanan2017simple} involves independently training multiple DNNs with different initialization, and the variance in the predictions of the multiple DNNs serves as an uncertainty measure. DE is robust and reliable in uncertainty estimation but computationally and memory-wise expensive, requiring training and storing multiple models. UQ methods have been applied in many domains, e.g., cyber-physical systems~\cite{xu2024pretrain,catak2022uncertainty}, computer vision tasks~\cite{su2023uncertainty,feng2018towards,kendall2017uncertainties}, and healthcare~\cite {lu2023evoclinical}. In contrast, this paper adopts MC-dropout as the UQ method to quantify the uncertainty of sticker detection tasks in laptop refurbishing robots, thereby studying their real-world application in a new context.

\textit{Robustness Assessment of DNNs.} Robustness measures DNNs' reliability. Recent robustness evaluation methods focus on adversarial robustness~\cite{carlini2019evaluating,brendel2019accurate,croce2020reliable}, which refers to the ability of DNNs to maintain performance and provide reliable outputs in the face of various perturbations, noises, and adversarial attacks. For instance, Carlini and Wagner~\cite{carlini2017towards} constructed three adversarial attack methods to evaluate the robustness of defensively distilled networks. Madry et al.~\cite{madry2017towards} proposed a Projected Gradient Descent (PGD) attack to assess the adversarial robustness of DNNs from the robust optimization perspective. 
This paper employs a dense adversarial attack technique to measure adversarial robustness. Regarding the robustness metrics, in addition to measuring the robustness using the performance metric (i.e., \map), we further calculate the robustness score concerning \uqms{s}.

\section{Conclusions and Future Work} \label{sec:conclusion}
In this paper, we conduct an empirical study to evaluate the detection accuracy, prediction uncertainty, and adversarial robustness of six sticker detection models in the laptop refurbishing software. We adopt the Monte Carlo Dropout method to quantify the prediction uncertainty and measure uncertainty from two aspects: uncertainty in classification and regression. Besides, we present novel robustness metrics to evaluate the robustness of the models regarding detection accuracy and uncertainty. 
The results show \fasterrcnnv and \retinanetv achieve the overall best performance regarding all metrics. 
Future works include quantifying uncertainties of the entire laptop refurbishing process and automatically generating realistic image datasets.

\section*{Acknowledgement}
The work is supported by the RoboSapiens project funded by the European Commission’s Horizon Europe programme under grant agreement number 101133807. Jiahui Wu is also partially supported by the Co-tester project from the Research Council of Norway (No. 314544).

\bibliographystyle{unsrt}  
\bibliography{references}

\begin{thebibliography}{10}

\bibitem{doi/10.2779/05068}
European Commission and Directorate-General for Communication.
\newblock {\em Circular economy action plan – For a cleaner and more competitive Europe}.
\newblock Publications Office of the European Union, 2020.

\bibitem{carlini2017towards}
Nicholas Carlini and David Wagner.
\newblock Towards evaluating the robustness of neural networks.
\newblock In {\em 2017 ieee symposium on security and privacy (sp)}, pages 39--57. Ieee, 2017.

\bibitem{gal2016dropout}
Yarin Gal and Zoubin Ghahramani.
\newblock Dropout as a bayesian approximation: Representing model uncertainty in deep learning.
\newblock In {\em international conference on machine learning}, pages 1050--1059. PMLR, 2016.

\bibitem{xie2017adversarial}
Cihang Xie, Jianyu Wang, Zhishuai Zhang, Yuyin Zhou, Lingxi Xie, and Alan Yuille.
\newblock Adversarial examples for semantic segmentation and object detection.
\newblock In {\em Proceedings of the IEEE international conference on computer vision}, pages 1369--1378, 2017.

\bibitem{betker2023improving}
James Betker, Gabriel Goh, Li~Jing, Tim Brooks, Jianfeng Wang, Linjie Li, Long Ouyang, Juntang Zhuang, Joyce Lee, Yufei Guo, et~al.
\newblock Improving image generation with better captions.
\newblock {\em Computer Science. https://cdn. openai. com/papers/dall-e-3. pdf}, 2(3):8, 2023.

\bibitem{rombach2021highresolution}
Robin Rombach, Andreas Blattmann, Dominik Lorenz, Patrick Esser, and Björn Ommer.
\newblock High-resolution image synthesis with latent diffusion models, 2021.

\bibitem{tran2019bayesian}
Dustin Tran, Mike Dusenberry, Mark Van Der~Wilk, and Danijar Hafner.
\newblock Bayesian layers: A module for neural network uncertainty.
\newblock {\em Advances in neural information processing systems}, 32, 2019.

\bibitem{mackay1992practical}
David~JC MacKay.
\newblock A practical bayesian framework for backpropagation networks.
\newblock {\em Neural computation}, 4(3):448--472, 1992.

\bibitem{srivastava2014dropout}
Nitish Srivastava, Geoffrey Hinton, Alex Krizhevsky, Ilya Sutskever, and Ruslan Salakhutdinov.
\newblock Dropout: a simple way to prevent neural networks from overfitting.
\newblock {\em The journal of machine learning research}, 15(1):1929--1958, 2014.

\bibitem{bernardo2009bayesian}
Jos{\'e}~M Bernardo and Adrian~FM Smith.
\newblock {\em Bayesian theory}, volume 405.
\newblock John Wiley \& Sons, 2009.

\bibitem{campello2013density}
Ricardo~JGB Campello, Davoud Moulavi, and J{\"o}rg Sander.
\newblock Density-based clustering based on hierarchical density estimates.
\newblock In {\em Pacific-Asia conference on knowledge discovery and data mining}, pages 160--172. Springer, 2013.

\bibitem{gal2016uncertainty}
Yarin Gal et~al.
\newblock Uncertainty in deep learning.
\newblock 2016.

\bibitem{freeman1965elementary}
Linton~C Freeman.
\newblock Elementary applied statistics: for students in behavioral science.
\newblock {\em (No Title)}, 1965.

\bibitem{shannon1948mathematical}
Claude~Elwood Shannon.
\newblock A mathematical theory of communication.
\newblock {\em The Bell system technical journal}, 27(3):379--423, 1948.

\bibitem{feng2018towards}
Di~Feng, Lars Rosenbaum, and Klaus Dietmayer.
\newblock Towards safe autonomous driving: Capture uncertainty in the deep neural network for lidar 3d vehicle detection.
\newblock In {\em 2018 21st international conference on intelligent transportation systems (ITSC)}, pages 3266--3273. IEEE, 2018.

\bibitem{catak2021prediction}
Ferhat~Ozgur Catak, Tao Yue, and Shaukat Ali.
\newblock Prediction surface uncertainty quantification in object detection models for autonomous driving.
\newblock In {\em 2021 IEEE International Conference on Artificial Intelligence Testing (AITest)}, pages 93--100. IEEE, 2021.

\bibitem{zhang2024vision}
Jingyi Zhang, Jiaxing Huang, Sheng Jin, and Shijian Lu.
\newblock Vision-language models for vision tasks: A survey.
\newblock {\em IEEE Transactions on Pattern Analysis and Machine Intelligence}, 2024.

\bibitem{li2023gligen}
Yuheng Li, Haotian Liu, Qingyang Wu, Fangzhou Mu, Jianwei Yang, Jianfeng Gao, Chunyuan Li, and Yong~Jae Lee.
\newblock Gligen: Open-set grounded text-to-image generation.
\newblock In {\em Proceedings of the IEEE/CVF Conference on Computer Vision and Pattern Recognition}, pages 22511--22521, 2023.

\bibitem{madry2017towards}
Aleksander Madry, Aleksandar Makelov, Ludwig Schmidt, Dimitris Tsipras, and Adrian Vladu.
\newblock Towards deep learning models resistant to adversarial attacks.
\newblock {\em arXiv preprint arXiv:1706.06083}, 2017.

\bibitem{ren2015faster}
Shaoqing Ren, Kaiming He, Ross Girshick, and Jian Sun.
\newblock Faster r-cnn: Towards real-time object detection with region proposal networks.
\newblock {\em Advances in neural information processing systems}, 28, 2015.

\bibitem{li2021benchmarking}
Yanghao Li, Saining Xie, Xinlei Chen, Piotr Dollar, Kaiming He, and Ross Girshick.
\newblock Benchmarking detection transfer learning with vision transformers.
\newblock {\em arXiv preprint arXiv:2111.11429}, 2021.

\bibitem{lin2017focal}
Tsung-Yi Lin, Priya Goyal, Ross Girshick, Kaiming He, and Piotr Doll{\'a}r.
\newblock Focal loss for dense object detection.
\newblock In {\em Proceedings of the IEEE international conference on computer vision}, pages 2980--2988, 2017.

\bibitem{zhang2020bridging}
Shifeng Zhang, Cheng Chi, Yongqiang Yao, Zhen Lei, and Stan~Z Li.
\newblock Bridging the gap between anchor-based and anchor-free detection via adaptive training sample selection.
\newblock In {\em Proceedings of the IEEE/CVF conference on computer vision and pattern recognition}, pages 9759--9768, 2020.

\bibitem{liu2016ssd}
Wei Liu, Dragomir Anguelov, Dumitru Erhan, Christian Szegedy, Scott Reed, Cheng-Yang Fu, and Alexander~C Berg.
\newblock Ssd: Single shot multibox detector.
\newblock In {\em Computer Vision--ECCV 2016: 14th European Conference, Amsterdam, The Netherlands, October 11--14, 2016, Proceedings, Part I 14}, pages 21--37. Springer, 2016.

\bibitem{howard2019searching}
Andrew Howard, Mark Sandler, Grace Chu, Liang-Chieh Chen, Bo~Chen, Mingxing Tan, Weijun Wang, Yukun Zhu, Ruoming Pang, Vijay Vasudevan, et~al.
\newblock Searching for mobilenetv3.
\newblock In {\em Proceedings of the IEEE/CVF international conference on computer vision}, pages 1314--1324, 2019.

\bibitem{paszke2019pytorch}
Adam Paszke, Sam Gross, Francisco Massa, Adam Lerer, James Bradbury, Gregory Chanan, Trevor Killeen, Zeming Lin, Natalia Gimelshein, Luca Antiga, et~al.
\newblock Pytorch: An imperative style, high-performance deep learning library.
\newblock {\em Advances in neural information processing systems}, 32, 2019.

\bibitem{everingham2010pascal}
Mark Everingham, Luc Van~Gool, Christopher~KI Williams, John Winn, and Andrew Zisserman.
\newblock The pascal visual object classes (voc) challenge.
\newblock {\em International journal of computer vision}, 88:303--338, 2010.

\bibitem{friedman1937use}
Milton Friedman.
\newblock The use of ranks to avoid the assumption of normality implicit in the analysis of variance.
\newblock {\em Journal of the american statistical association}, 32(200):675--701, 1937.

\bibitem{wilcoxon1992individual}
Frank Wilcoxon.
\newblock Individual comparisons by ranking methods.
\newblock In {\em Breakthroughs in statistics: Methodology and distribution}, pages 196--202. Springer, 1992.

\bibitem{kerby2014simple}
Dave~S Kerby.
\newblock The simple difference formula: An approach to teaching nonparametric correlation.
\newblock {\em Comprehensive Psychology}, 3:11--IT, 2014.

\bibitem{mangiafico2016summary}
Salvatore~S Mangiafico.
\newblock Summary and analysis of extension program evaluation in r.
\newblock {\em Rutgers Cooperative Extension: New Brunswick, NJ, USA}, 125:16--22, 2016.

\bibitem{garcia2010advanced}
Salvador Garc{\'\i}a, Alberto Fern{\'a}ndez, Juli{\'a}n Luengo, and Francisco Herrera.
\newblock Advanced nonparametric tests for multiple comparisons in the design of experiments in computational intelligence and data mining: Experimental analysis of power.
\newblock {\em Information sciences}, 180(10):2044--2064, 2010.

\bibitem{tukey1977exploratory}
John~Wilder Tukey et~al.
\newblock {\em Exploratory data analysis}, volume~2.
\newblock Springer, 1977.

\bibitem{reimann2005background}
Clemens Reimann, Peter Filzmoser, and Robert~G Garrett.
\newblock Background and threshold: critical comparison of methods of determination.
\newblock {\em Science of the total environment}, 346(1-3):1--16, 2005.

\bibitem{spearman1961proof}
Charles Spearman.
\newblock The proof and measurement of association between two things.
\newblock 1961.

\bibitem{holm1979simple}
Sture Holm.
\newblock A simple sequentially rejective multiple test procedure.
\newblock {\em Scandinavian journal of statistics}, pages 65--70, 1979.

\bibitem{github}
ComplexSE.
\newblock Assessing the uncertainty and robustness of object detection models for detecting stickers on laptops.
\newblock \url{https://github.com/Simula-COMPLEX/laptop-sticker-uncertainty-robustness}.

\bibitem{wu2024reality}
Jiahui Wu, Chengjie Lu, Aitor Arrieta, Tao Yue, and Shaukat Ali.
\newblock Reality bites: Assessing the realism of driving scenarios with large language models.
\newblock In {\em Proceedings of the 2024 IEEE/ACM First International Conference on AI Foundation Models and Software Engineering}, pages 40--51, 2024.

\bibitem{lakshminarayanan2017simple}
Balaji Lakshminarayanan, Alexander Pritzel, and Charles Blundell.
\newblock Simple and scalable predictive uncertainty estimation using deep ensembles.
\newblock {\em Advances in neural information processing systems}, 30, 2017.

\bibitem{xu2024pretrain}
Qinghua Xu, Tao Yue, Shaukat Ali, and Maite Arratibel.
\newblock Pretrain, prompt, and transfer: Evolving digital twins for time-to-event analysis in cyber-physical systems.
\newblock {\em IEEE Transactions on Software Engineering}, 2024.

\bibitem{catak2022uncertainty}
Ferhat~Ozgur Catak, Tao Yue, and Shaukat Ali.
\newblock Uncertainty-aware prediction validator in deep learning models for cyber-physical system data.
\newblock {\em ACM Transactions on Software Engineering and Methodology (TOSEM)}, 31(4):1--31, 2022.

\bibitem{su2023uncertainty}
Sanbao Su, Yiming Li, Sihong He, Songyang Han, Chen Feng, Caiwen Ding, and Fei Miao.
\newblock Uncertainty quantification of collaborative detection for self-driving.
\newblock In {\em 2023 IEEE International Conference on Robotics and Automation (ICRA)}, pages 5588--5594. IEEE, 2023.

\bibitem{kendall2017uncertainties}
Alex Kendall and Yarin Gal.
\newblock What uncertainties do we need in bayesian deep learning for computer vision?
\newblock {\em Advances in neural information processing systems}, 30, 2017.

\bibitem{lu2023evoclinical}
Chengjie Lu, Qinghua Xu, Tao Yue, Shaukat Ali, Thomas Schwitalla, and Jan~F Nyg{\aa}rd.
\newblock Evoclinical: Evolving cyber-cyber digital twin with active transfer learning for automated cancer registry system.
\newblock In {\em Proceedings of the 31st ACM Joint European Software Engineering Conference and Symposium on the Foundations of Software Engineering}, pages 1973--1984, 2023.

\bibitem{carlini2019evaluating}
Nicholas Carlini, Anish Athalye, Nicolas Papernot, Wieland Brendel, Jonas Rauber, Dimitris Tsipras, Ian Goodfellow, Aleksander Madry, and Alexey Kurakin.
\newblock On evaluating adversarial robustness.
\newblock {\em arXiv preprint arXiv:1902.06705}, 2019.

\bibitem{brendel2019accurate}
Wieland Brendel, Jonas Rauber, Matthias K{\"u}mmerer, Ivan Ustyuzhaninov, and Matthias Bethge.
\newblock Accurate, reliable and fast robustness evaluation.
\newblock {\em Advances in neural information processing systems}, 32, 2019.

\bibitem{croce2020reliable}
Francesco Croce and Matthias Hein.
\newblock Reliable evaluation of adversarial robustness with an ensemble of diverse parameter-free attacks.
\newblock In {\em International conference on machine learning}, pages 2206--2216. PMLR, 2020.

\end{thebibliography}

\end{document}